\documentclass[runningheads]{article}

\usepackage[affil-it]{authblk}
\usepackage[utf8]{inputenc}
\usepackage{graphicx}
\usepackage{algorithm,algorithmicx,algpseudocode}
\usepackage{amsmath,amssymb}
\usepackage{bbm}
\usepackage{hyperref}
\usepackage{url}
\usepackage{xcolor}
\usepackage{multirow}
\usepackage{array}
\usepackage[export]{adjustbox}

\newcommand{\median}[1]{\bar{#1}}

\newcolumntype{M}[1]{>{\centering\arraybackslash}m{#1}}
\newcolumntype{N}{@{}m{0pt}@{}}

\begin{document}

\title{Generalized Median Graph via Iterative Alternate Minimizations}

\author{Nicolas Boria$^\dagger$ and 
S{\'e}bastien Bougleux$^\ddagger$ and 
Benoit Ga{\"u}z{\`e}re$^*$ and 
Luc Brun$^\dagger$}
%
\affil{
  $\dagger$ Normandie Univ, ENSICAEN, UNICAEN, CNRS, GREYC, Caen, France\\
  $\ddagger$ Normandie Univ, UNICAEN, ENSICAEN, CNRS, GREYC, Caen, France\\
$*$ Normandie Univ, INSA ROUEN Normandie, LITIS, Rouen, France}
%

\maketitle

\begin{abstract}
Computing a graph prototype may constitute a core element for clustering or classification tasks. However, its computation is an NP-Hard problem, even for simple classes of graphs. In this paper, we propose an efficient approach based on block coordinate descent to compute a generalized median graph from a set of graphs. This approach relies on a clear definition of the optimization process and handles labeling on both edges and nodes. This iterative process optimizes the edit operations to perform on a graph alternatively on nodes and edges.
Several experiments on different datasets show the efficiency of our approach.

\end{abstract}

\section{Introduction}
In a wide variety of scientific domains, attributed graphs provide a powerful structure to represent, process and analyze data. However, determining fundamental tools such as a distance or an average graph is non trivial. Given a space $\mathbb{G}$ of attributed graphs, \textit{Graph Edit Distance} (GED) is a natural choice for comparing graphs \cite{bunke:1983aa,riebook}. It measures the minimal amount of distortion needed to transform a graph into another by means of edit operations. It can be defined as a minimal-path problem which relies on a cost function acting as a metric in $\mathbb{G}$, and rewritten as a special quadratic assignment problem close to the graph matching problem. Computing Graph Edit Distance is NP-Hard and still cannot be solved in a reasonable time for graphs exceeding a dozen of vertices, even for simple cost functions. Therefore, several strategies have been explored to provide tight upper-bounds in polynomial time \cite{riebook}. Computing a representative of a set of graphs $\mathcal{G}\subset\mathbb{G}$ is even more difficult. It commonly consists in finding a \textit{generalized median graph}, \textit{ie.} a graph $\median{G}\in\mathbb{G}$ that minimizes the \textit{sum of distances} (SOD) to all the graphs in $\mathcal{G}$~\cite{jiang01}:
\begin{equation}\label{eq-gmg}
    \median{G} \in \arg \min_{G\in\mathbb{G}}\sum_{G'\in \mathcal{G}}d(G,G')
\end{equation}
where $d:\mathbb{G}\times\mathbb{G}\rightarrow\mathbb{R}_+$ denotes Graph Edit Distance.
Exact methods are restricted to labeled graphs with particular cost functions or datasets containing a small total number of vertices \cite{ferrer-pdh}. To estimate median graphs in a reasonable computational time, several methods reduce the SOD by a local search around an initial candidate graph, by genetic search \cite{jiang01}, greedy search based on partitioning vertices of different graphs \cite{Hlaoui2006}, greedy adaptive search \cite{MUSMANNO2016}, or linearization and discrete optimization \cite{Mukherjee2009}. A different strategy is based on graph embedding \cite{FERRER2010,FERRER2011,Ferrer2013,nienk16,CHAIEB2017}, usually with distances between graphs as coordinates. A representative is more easily computed within this space. Then a median graph is reconstructed by going back to the original space of graphs. While these approaches are able to tackle the complexity of the previous ones, the link with the definition of a generalized median graph is not trivial and difficult to analyze. Other approaches use the relationship between common-labeling and the median graph to derive bounds on the SOD \cite{rebagliati12}, or extend the concept of representative to correspondences between graphs~\cite{MORENOGARCIA2018}.

In this paper, we propose to estimate a generalized median graph by a block coordinate descent that iterates two minimization steps from an initial candidate (Sec.~\ref{sec:estimate-gm}): one for updating the SOD w.r.t. edges and attributes on nodes and on edges, and the other w.r.t. distances. The order of the resulting graph is fixed before the descent process by the order of the initial candidate. This candidate is set to  a \textit{set-median}, \textit{i.e.} a graph of $\mathcal{G}$ minimizing the SOD ($\mathbb{G}$ restricted to $\mathcal{G}$ in Eq.~\ref{eq-gmg}). While the first step of the descent shares similarities with the update presented in \cite{jiang01}, the update rules are not the same, and any algorithm can be used to estimate GED in the second step or for initialization. The first empirical results on two datasets (Sec.~\ref{sec:xp}) show on the one hand that the proposed method systematically reduces the SOD associated with the initial candidate, \textit{i.e.} a set-median, and on the other hand that the accuracy of the approximate GED has more impact on the descent than on the computation of a set-median. The following section introduces the expressions we use to facilitate the derivation of the proposed algorithm.

\section{Graph Transformations and Graph Edit Distance}\label{sec:preambule}
\begin{figure}[!t]
\includegraphics[scale=0.7,valign=c]{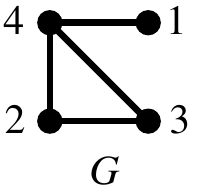}
$\begin{array}{c}
\varphi=(1,2,2,3)\\
\Phi=\left(\begin{matrix}0&0&0&3\\0&0&1&2\\0&1&0&3\\3&2&3&0\end{matrix}\right)
\end{array}$
\quad\includegraphics[scale=0.7,valign=c]{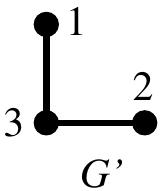}
$\begin{array}{c}
\varphi'=(1,2,2)\\
\Phi'=\left(\begin{matrix}0&0&4\\0&0&1\\4&1&0\end{matrix}\right)
\end{array}$
\quad\includegraphics[scale=0.7,valign=c]{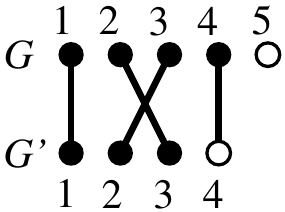}
$\begin{array}{c}
\pi=(1,3,2,4)\\
\pi'=(1,3,2)
\end{array}$
\caption{\label{fig-s2}Labeled graphs (label $0$ if no edge) and a transformation $(\pi,\pi')$ of their vertices. Induced operations on edges: $\phi_{2,3}=\phi_{3,2}$ substituted by $\phi_{3,2}'=\phi_{2,3}'$, $(1,4),(2,4),(3,4)$ removed from $G$, $(1,2)$ inserted in $G$ from $(1,3)$ in $G'$ with $\phi_{1,2}=\phi_{2,1}=\phi_{1,3}'$.}
\end{figure}

We consider simple undirected attributed graphs. An attributed graph $G$ of order $n$ can be  encoded by a triplet $(\varphi,A,\Phi)$ (Fig.~\ref{fig-s2}). The $n$-tuple $\varphi=(\varphi_i)_{i}$ associates an attribute (or feature) $\varphi_i$ of a space $\mathbb{F}_v$ to each integer $i\in[n]=\{1,\ldots,n\}$ (vertices are represented by the set $[n]$). $A\in\{0,1\}^{n\times n}$ is the vertex-vertex adjacency matrix of $G$, \textit{i.e.} $a_{i,j}=1$ if there is an edge $(i,j)$, else $a_{i,j}=0$. $\Phi=(\phi_{i,j})_{i,j}$ associates an attribute $\phi_{i,j}$ of a space $\mathbb{F}_e$ to each pair $(i,j)\in[n]\times[n]$. When $(i,j)$ is not an edge, $\phi_{i,j}$ can be equal to any value, it does not affect the following expressions. Obviously, $A$ and $\Phi$ are symmetric. Let $\mathbb{G}$ be the space of all attributed graphs for $\mathbb{F}_v$ and $\mathbb{F}_e$ fixed. In this paper, each space of attributes is restricted to a finite set of positive integer labels, or to the Euclidean space.

A graph $G=(\varphi,A,\Phi)$ of order $n$ can be transformed into a graph $G'=(\varphi',A',\Phi')$ of order $n'$ by applying a composition of elementary transformations, a.k.a. \textit{edit operations}, to $G$. An edit operation transforms a graph into another by either removing an element (a vertex or an edge), substituting an attribute attached to an element by another attribute, or by inserting an element and its attribute (between two existing vertices for edges). Moreover, if each element of both graphs is assumed to be involved in exactly one edit operation, the number of operations is minimized, and the transformation of $G$ into $G'$ is fully described by the transformation of the vertices of $G$ into the ones of $G'$. Here, this transformation, a.k.a. \textit{error-correcting matching} \cite{bunke:1983aa,riebook}, is defined as a pair $(\pi,\pi')\in[n'+1]^n\times[n+1]^{n'}$ so that $\pi_i=k\in[n']\Leftrightarrow\pi'_k=i\in[n]$ (Fig.~\ref{fig-s2}). Each vertex $i$ of $G$ is either substituted by a vertex $k$ of $G'$ ($\pi_i=k$ and $\pi'_k=i$), or removed ($\pi_i=n'+1$). Each vertex $k$ of $G'$ that is not substituted to a vertex of $G$ is inserted ($\pi'_k=n+1$). The transformation of the edges of $G$ into the ones of $G'$ is induced by the transformation of the vertices. The set $\{(i,j)\in[n]\times[n]\,|\,a_{i,j}=1\wedge\pi_i\in[n']\wedge\pi_j\in[n']\wedge a_{\pi_i,\pi_j}=1\}$ defines the substituted edges, the set $\{(i,j)\in[n]\times[n]\,|\,a_{i,j}=1\wedge((\pi_i\in[n']\wedge\pi_j\in[n']\wedge a_{\pi_i,\pi_j}=0)\vee\pi_i=n'+1\vee\pi_j=n'+1)\}$ defines the removed edges, and the set $\{(k,l)\in[n']\times[n']\,|\,a'_{k,l}=1\wedge((\pi'_k\in[n]\wedge\pi'_l\in[n]\wedge a_{\pi'_k,\pi'_l}=0)\vee\pi'_k=n+1\vee\pi'_l=n+1)\}$ defines the inserted edges. Since $\pi'$ can be obtained from $\pi$, we omit $\pi'$ for simplicity, and we denote by $\Pi(G,G')$ all the transformations of $G$ to $G'$.

A transformation $\pi^\star\in\Pi(G,G')$ is said to be minimal if its cost is minimal, \textit{i.e.} if $c(\pi^\star,G,G')=\min_{\pi\in\Pi(G,G')}c(\pi,G,G')$, with $c(\pi,G,G')=c_v(\pi,\varphi,\varphi')+\tfrac{1}{2}c_e(\pi,A,\Phi,A',\Phi')$ the cost for transforming $G$ into $G'$ using $\pi$, and
\begin{align}
    \label{eq:dv}&c_v(\pi,\varphi,\varphi')=\sum_{i=1}^{n}\delta_{\pi_i}\,c_{\text{vfs}}(\varphi_i,\varphi'_{\pi_i})+(1-\delta_{\pi_i})c_{\text{vr}}+\sum_{k=1}^{n'}(1-\delta_{\pi'_k})\,c_{\text{vi}}\\
    \label{eq:de}&\begin{array}{l}
    c_e(\pi,A,\Phi,A',\Phi')=\sum_{i=1}^{n}\sum_{j=1}^{n}\delta_{\pi_i\pi_j}\,a_{i,j}\,a'_{\pi_i,\pi_j}\,c_{\text{efs}}\left(\phi_{i,j},\phi'_{\pi_i\pi_j}\right)\\
    \qquad~ +c_{\text{er}}\sum_{i=1}^{n}\sum_{j=1}^{n}\delta_{\pi_i\pi_j}a_{i,j}(1-a'_{\pi_i,\pi_j})+(1-\delta_{\pi_i\pi_j})a_{i,j}\\
    \qquad~ +c_{\text{ei}}\sum_{i=1}^{n}\sum_{j=1}^{n}\delta_{\pi_i\pi_j}(1-a_{i,j})a'_{\pi_i\pi_j}+c_{\text{ei}}\sum_{k=1}^{n'}\sum_{l=1}^{n'}(1-\delta_{\pi'_k\pi'_l})a'_{k,l}
\end{array}
\end{align}
the costs for transforming attributed vertices and edges, respectively. $\delta_{\pi_i}=1$ if $\pi_i\in[n']$, else $0$, and $\delta_{\pi_i\pi_j}=\delta_{\pi_i}\delta_{\pi_j}$. Functions $c_{\text{vfs}}:\mathbb{F}_v\times\mathbb{F}_v\rightarrow[0,+\infty)$ and $c_{\text{efs}}:\mathbb{F}_e\times\mathbb{F}_e\rightarrow[0,+\infty)$ measure costs to substitute vertices and edges. In this paper, the costs for removing and inserting elements are restricted to positive constants, denoted $c_{\text{vr}}$, $c_{\text{vi}}$, $c_{\text{er}}$, $c_{\text{ei}}$. When any substitution of elements is no more expensive than removing and inserting these elements, \textit{Graph Edit Distance} (GED) between $G$ and $G'$ is equal to the cost of a minimal transformation \cite{riebook}: $d(G,G')=\min_{\pi\in\Pi(G,G')}\,c(\pi,G,G')$. This case is considered in the sequel.
\section{Estimating a Generalized Median Graph}
\label{sec:estimate-gm}
Given a set of graphs $\mathcal{G}=\{G_p\}_p\subset\mathbb{G}$, with $G_p=(\varphi_p,A_p,\phi_p)$ of order $n_p$, a \textit{generalized median graph} $\bar{G}=(\bar{\varphi},\bar{A},\bar{\phi})\in\mathbb{G}$ of $\mathcal{G}$ minimizes the \textit{sum of distances} (SOD) to the graphs of $\mathcal{G}$ \cite{jiang01,ferrer-pdh}: $s(\bar{G},\mathcal{G})=\min_{G\in\mathbb{G}}\,s(G,\mathcal{G})$, with
	$s(G,\mathcal{G})=\sum_{G_p\in\mathcal{G}}d(G,G_p)=\sum_{p=1}^{|\mathcal{G}|}\min_{\pi_p\in\Pi(G,G_p)}c(\pi_p,G,G_p)$. 
We propose to use a block coordinate descent to estimate both $\bar{G}$ and the minimal transformations $(\pi_p)_p$.
\subsection{Proposed algorithm}\label{sec:algo}
First, $\bar{G}$ is initialized to a set-median of $\mathcal{G}$, \textit{i.e.}  $\bar{G}=\arg\min_{G_p\in\mathcal{G}}s(G_p,\mathcal{G})$. It can be computed in $O(a|\mathcal{G}|^2)$ time \cite{ferrer-pdh}, where $a$ is the complexity of the algorithm used for computing or estimating GED. This also provides the minimal transformations $(\bar{\pi}_p)_p$ from $\bar{G}$ to the graphs of $\mathcal{G}$. The order $\bar{n}$ of $\bar{G}$ is then fixed, \textit{i.e.} considered as a constant in the optimization process. Then, $(\bar{\varphi},\bar{A},\bar{\Phi})$ and $(\bar{\pi}_p)_p$ are alternatively updated as follows:
\begin{align}
    &\bar{G}=(\bar{\varphi},\bar{A},\bar{\Phi})\leftarrow\arg\min\limits_{\substack{\varphi\in\mathbb{F}_v^{\bar{n}}\\A\in\{0,1\}^{\bar{n}\times\bar{n}}\\\Phi\in\mathbb{F}_e^{\bar{n}\times\bar{n}}}}\,\sum_{p=1}^{|\mathcal{G}|}c_v(\bar{\pi}_p,\varphi,\varphi_p)+\tfrac{1}{2}c_e(\bar{\pi}_p,A,\Phi,A_p,\Phi_p) \label{eq-find-gmg}\\
    &\forall p\in\{1,\ldots,|\mathcal{G}|\},\quad \bar{\pi}_p\leftarrow\arg\min\limits_{\pi_p\in\Pi(\bar{G},G_p)}c(\pi_p,\bar{G},G_p)\label{eq-find-ged}
\end{align}
until convergence, that is, until a stability is reached both in $\bar{G}$ and $(\bar{\pi}_p)_p$. The resolution of the minimization of the sum of distances when the transformations are fixed (Eq.~\ref{eq-find-gmg}) mainly depends on the nature of $\mathbb{F}_v$ and $\mathbb{F}_e$, as well as the form of the cost functions $c_{\text{vfs}}$ and $c_{\text{vef}}$. This is detailed later in this section, in particular it can be solved in $O(\bar{n}^2|\mathcal{G}|)$ time under some conditions. The update of the transformations (Eq.~\ref{eq-find-ged}) consists in solving $|\mathcal{G}|$ times GED problem, so in $O(a|\mathcal{G}|)$ time. Since the order $\bar{n}$ is fixed, and GED can usually be only estimated, the algorithm may not converge to the true generalized median graph.

We assume that an algorithm for computing GED is given, and we focus on the minimization of the sum of distances w.r.t. the graph (Eq.~\ref{eq-find-gmg}). It can be decomposed into two independent minimizations as long as the attributes $\varphi_p$ and $\Phi_p$ are independent for each $p$, that we consider in this paper:
\begin{equation}\label{eq:svse}
    \bar{\varphi}\leftarrow\arg\min_{\varphi\in\mathbb{F}_v^{\bar{n}}}s_v(\varphi),\quad(\bar{A},\bar{\Phi})\leftarrow\arg\min_{\substack{A\in\{0,1\}^{\bar{n}\times\bar{n}}\\\Phi\in\mathbb{F}_e^{\bar{n}\times\bar{n}}}}s_e(A,\phi)
\end{equation}
with $s_v(\varphi)=\sum_{p=1}^{|\mathcal{G}|}c_v(\bar{\pi}_p,\varphi,\varphi_p)$ and $s_e(\phi,A)=\sum_{p=1}^{|\mathcal{G}|}c_e(\bar{\pi}_p,A,\phi,A_p,\phi_p)$. The minimization of each term is detailed in the two following sections. Note that some results are already presented in \cite{jiang01}, in particular for vertices. There are obtained in a different way, allowing to take into account more easily different spaces of attributes and cost functions associated to edit operations.
\subsection{Updating vertex attributes}
Only the cost function $c_{\text{vfs}}$ depends on vertex attributes in the expression of $c_v$ (Eq.~\ref{eq:dv}). So the attributes $\bar{\varphi}$ in Eq.~\ref{eq:svse} are updated by solving the equivalent problem 
$
    \arg\min_{\varphi\in\mathbb{F}_v^{\bar{n}}}\sum_{i=1}^{\bar{n}}f_i(\varphi_i),
$
with the function $f_i:\mathbb{F}_v\rightarrow\mathbb{R}_+$ defined by $f_i(\varphi_i)\,{=}\,\sum_{p=1}^{|\mathcal{G}|}\delta_{\pi^p_i}\,c_{\text{vfs}}(\varphi_i,\varphi^p_{\pi^p_i})$. The objective function is a sum of positive and independent terms $f_i$, so the attributes are updated by:
\begin{equation}\label{eq:optvert}
	\forall i=1,\ldots,\bar{n},\quad\bar{\varphi}_i\leftarrow\arg\min_{\varphi_i\in\mathbb{F}_v}f_i(\varphi_i)
\end{equation}
The solution depends on $\mathbb{F}_v$ and on the cost function $c_{\text{vfs}}$.

When attributes are labels ($\mathbb{F}_v\subset\mathbb{N}$), the cost for substituting a label $x\in\mathbb{F}_v$ by a label $y\in\mathbb{F}_v$ is defined as $c_{\text{vfs}}(x,y)=c_{\text{vs}}(1-\delta_{x,y})$, with $c_{\text{vs}}>0$ a constant, \textit{i.e.} $0$ if the labels are the same, and $c_{\text{vs}}$ otherwise. Then $f_i$ can be rewritten as $f_i(\varphi_i) = \sum_{p=1}^{|\mathcal{G}|}\delta_{\pi^p_i}\,c_{\text{vs}}(1-\delta_{\varphi_i,\varphi^p_{\pi^p_i}})= c_{\text{vs}}(|S_i|-h_i^0(\varphi_i))$, 
where $S_i=\{\pi^p_i\,|\,\pi^p_i\in[n_p],\,p=1,\ldots,|\mathcal{G}|\}$ is the set of vertices that are substituted to $i$ by the mappings $\pi_p$, and $h_i^0:\mathbb{F}_v\rightarrow\{0,\ldots,|\mathcal{G}|\}\subset\mathbb{N}$,
$
    h_i^0(\varphi_i)=\sum_{p=1}^{|\mathcal{G}|}\delta_{\pi^p_i}\delta_{\varphi_i,\varphi^p_{\pi^p_i}}
$,
counts the number of times $i$ is substituted by a vertex having the same label (with zero cost). So the attributes (Eq.~\ref{eq:optvert}) are updated by:
\begin{equation}\label{eq:labvert}
	\forall i=1,\ldots,\bar{n},\quad\bar{\varphi}_i\leftarrow\arg\max_{\varphi_i\in\mathbb{F}_v}h^0_i(\varphi_i)
\end{equation}
Notice that $h_i^0$ can be pre-computed in $O(|\mathcal{G}|)$ time for each label of $\mathbb{F}_v$. The labels are thus updated for all the vertices of $\bar{G}$ in $O(\bar{n}|\mathcal{G}|)$ time at each iteration.

When $\mathbb{F}_v=\mathbb{R}^m$ is equipped with the scalar product $x^Ty=\sum_{k=1}^mx_ky_k$ and the $l_2$-norm $\|x\|=\sqrt{x^Tx}$, the cost for substituting an attribute $x$ by an attribute $y$ is defined by $c_{\text{vfs}}(x,y)=\|x-y\|^2$. In this case, we have:
$f_i(\varphi_i) = \sum_{p=1}^{|\mathcal{G}|}\delta_{\pi^p_i}\|\varphi_i-\varphi^p_{\pi^p_i}\|^2
$.
Any attribute $\bar{\varphi}_i$ satisfies $\nabla f_i(\bar{\varphi}_i)=0$, \textit{i.e.} $2\sum_p\delta_{\pi^p_i}(\bar{\varphi}_i-\varphi^p_{\pi^p_i})=0$, or:
\begin{equation}\label{eq:eucvert}
\forall i=1,\ldots,\bar{n},\quad\bar{\varphi}_i\leftarrow\frac{1}{\sum_{p=1}^{|\mathcal{G}|}\delta_{\pi^p_i}}\,\sum_{p=1}^{|\mathcal{G}|}\delta_{\pi^p_i}\,\varphi^{p}_{\pi^p_i}=\frac{1}{|S_i|}\sum_{p\in S_i}\varphi^p_{\pi^p_i}
\end{equation}
In other words, the optimal attribute for a vertex $i$ is given by the mean attribute of the vertices substituted to $i$ (the set $S_i$ defined in the previous paragraph). Once more, updating all the attributes is done in $O(\bar{n}|\mathcal{G}|)$ time at each iteration.
\subsection{Updating edges and their attributes}
The edges of $\bar{G}$, and their attributes, are computed at each step of the descent (Eq.~\ref{eq-find-gmg}) by minimizing $s_e$ (Eq.~\ref{eq:svse}). By removing the constant terms in $s_e$, \textit{i.e.} in $c_e$ (Eq.~\ref{eq:de}), it is easy to show that the minimization of $s_e$ can be rewritten as:
\begin{equation}\label{eq:optedge}
	\arg\min_{\substack{A\in\{0,1\}^{\bar{n}\times\bar{n}}\\\Phi\in\mathbb{F}_e^{\bar{n}\times\bar{n}}}}s_e(A,\phi)=\arg\min_{\substack{A\in\{0,1\}^{\bar{n}\times\bar{n}}\\\Phi\in\mathbb{F}_e^{\bar{n}\times\bar{n}}}}\sum_{i=1}^{\bar{n}}\sum_{j=1}^{\bar{n}}f_{i,j}(a_{i,j},\phi_{i,j})
\end{equation}
with the function $f_{i,j}:\{0,1\}\times\mathbb{F}_e\rightarrow\mathbb{R}_+$ defined by:
\begin{equation}\label{eq:fij}
\begin{array}{rl}
    f_{i,j}(a_{i,j},\phi_{i,j})=&a_{i,j}\sum_{p=1}^{|\mathcal{G}|}\delta_{\pi^p_i\pi^p_j}\,a^p_{\pi^p_i,\pi^p_j}\,c_{\text{efs}}(\phi_{i,j},\phi^p_{\pi^p_i,\pi^p_j})\\
    &+\,c_{\text{er}}a_{i,j}\sum_{p=1}^{|\mathcal{G}|}1-\delta_{\pi^p_i\pi^p_j}\,a^p_{\pi^p_i,\pi^p_j}\\
    &+\,c_{\text{ei}}(1-a_{i,j})\sum_{p=1}^{|\mathcal{G}|}\delta_{\pi^p_i\pi^p_j}a^p_{\pi^p_i,\pi^p_j}\\
    =&a_{i,j}\sum_{p=1}^{|\mathcal{G}|}\delta_{\pi^p_i\pi^p_j}\,a^p_{\pi^p_i,\pi^p_j}\,c_{\text{efs}}(\phi_{i,j},\phi^p_{\pi^p_i,\pi^p_j})\\
    &+\,c_{\text{er}}a_{i,j}\left(|\mathcal{G}|-|S_{i,j}|\right)\,+\,c_{\text{ei}}(1-a_{i,j})|S_{i,j}|
    \end{array}
\end{equation}
where $S_{i,j}=\{(\pi^p_i,\pi^p_j)\,|\,\pi^p_i\in[n_p]\wedge\pi^p_j\in[n_p]\wedge a^p_{\pi^p_i,\pi^p_j}=1,\,p=1,\ldots,|\mathcal{G}|\}$ is the set of edges that are substituted to $(i,j)$ by the mappings $\pi_p$. The terms $f_{i,j}$ are positive and independent from each others, so Eq.~\ref{eq:optedge} is equivalent to:
\begin{equation}\label{eq:optphia}
	\forall (i,j)\in[\bar{n}]\times[\bar{n}],\,i\not=j,\quad(\bar{a}_{i,j},\bar{\phi}_{i,j})\leftarrow\arg\min_{\substack{a_{i,j}\in\{0,1\}\\\phi_{i,j}\in\mathbb{F}_e}}f_{i,j}(a_{i,j},\phi_{i,j})
\end{equation}
Since $a_{i,j}$ can only take two values, if $a_{i,j}=0$ (no edge) then $f_{i,j}(0,\phi_{i,j})=c_{\text{ei}}|S_{i,j}|$ for any $\phi_{i,j}\in\mathbb{F}_e$, and if $a_{i,j}=1$ then $f_{i,j}(1,\phi_{i,j})$ is minimized for any
\begin{equation}\label{eq:optphiij}
	\phi_{i,j}^{\star}\in\arg\min_{\phi_{i,j}\in\mathbb{F}_e}\sum_{p=1}^{|\mathcal{G}|}\delta_{\pi^p_i\pi^p_j}\,a^p_{\pi^p_i,\pi^p_j}\,c_{\text{efs}}(\phi_{i,j},\phi^p_{\pi^p_i,\pi^p_j})
\end{equation}
By consequence $f_{i,j}$ is minimized for $\bar{\phi}_{i,j}=\phi_{i,j}^\star$ and
\begin{equation}\label{eq-optaij}
	\bar{a}_{i,j}=\left\lbrace\begin{array}{ll}
		1 & ~\text{if }f_{i,j}(1,\bar{\phi}_{ij})<c_{\text{ei}}|S_{i,j}|\\
		0 & ~\text{else }
	\end{array}\right.
\end{equation}
Solutions are finally obtained by solving Eq.~\ref{eq:optphiij}. It depends on $\mathbb{F}_e$ and $c_{\text{efs}}$.

When $\mathbb{F}_v\subset\mathbb{N}$ and $c_{\text{efs}}(x,y)=c_{\text{es}}(1-\delta_{x,y})$, with $c_{\text{es}}>0$ a constant, is the classical cost for labels, then $f_{i,j}$ (Eq.~\ref{eq:fij}) becomes 
\begin{equation*}
    f_{i,j}(a_{i,j},\phi_{i,j})=a_{i,j}\left(c_{\text{es}}\left(|S_{i,j}|-h_{i,j}^0(\phi_{i,j})\right)+c_{\text{er}}\left(|\mathcal{G}|-|S_{i,j}|\right)\right)+(1-a_{i,j})c_{\text{ei}}|S_{i,j}|
\end{equation*}
where $h^0_{i,j}(x)=\sum_{p=1}^{|\mathcal{G}|}\delta_{\pi^p_i\pi^p_j}a^p_{\pi^p_i,\pi^p_j}\delta_{x,\phi_{\pi^p_i,\pi^p_j}}$ counts the number of times $(i,j)$ is substituted by an edge having the label $x$. Then $\bar{\Phi}$ and $\bar{A}$ are updated for all $(i,j)\in[\bar{n}]\times[\bar{n}]$ by:
\begin{equation}\label{eq:optphiijlab}
\bar{\phi}_{i,j}\leftarrow\arg\max_{x\in\mathbb{F}_e}\,h^0_{i,j}(x)
\end{equation}
and
\begin{equation}\label{eq:optaijlab}
   \bar{a}_{i,j}\leftarrow\left\lbrace\begin{array}{ll}
         1 & ~\,\text{if}~\,h^0_{i,j}(\bar{\phi}_{i,j})>|\mathcal{G}|\frac{c_{\text{er}}}{c_{\text{es}}}+|S_{i,j}|\left(1-\frac{c_{\text{er}}+c_{\text{ei}}}{c_{\text{es}}}\right)\\
         0 & ~\,\text{else}
    \end{array}\right.
\end{equation}
Each edge $(i,j)$ is thus labeled with one of the most present labels among the ones substituted to $(i,j)$. Notice that $h_{i,j}^0:\mathbb{F}_e\rightarrow\{0,\ldots,|\mathcal{G}|\}$ and $|S_{i,j}|$ can be computed in $O(|\mathcal{G}|)$ time. So $\bar{\Phi}$ and $\bar{A}$ are computed in $O(\bar{n}^2|\mathcal{G}|)$ time.

Unlabeled graphs can be considered as labeled with a unique label, \textit{e.g.} $\mathbb{F}_e=\{1\}$. In this case $c_{\text{efs}}=0$ and $h_{i,j}^0=|S_{i,j}|$, so from Eq.~\ref{eq:optaijlab} $\bar{A}$ can be computed in $O(\bar{n}^2|\mathcal{G}|)$ time by:
\begin{equation}\label{eq:optaijunlab}
   \bar{a}_{i,j}\leftarrow\left\lbrace\begin{array}{ll}
         1 & ~\,\text{if}~\,|S_{i,j}|>|\mathcal{G}|\frac{c_{\text{er}}}{c_{\text{er}}+c_{\text{ei}}}\\
         0 & ~\,\text{else}
    \end{array}\right.
\end{equation}
\paragraph*{Remark.} Similar results can be derived for directed graphs, other spaces of attributes and other cost functions, for both vertices and edges. Due to limited space, it is restricted here to the cases considered in the experiments.

\section{Experimental results}\label{sec:xp}
In order to evaluate the validity of our method, the algorithm was implemented in \verb!C++! and tested on the datasets Letter (HIGH) \cite{riebook} and Monoterpenoides\,\footnote{GREYC Chemistry dataset: \url{https://brunl01.users.greyc.fr/CHEMISTRY/}}, a chemical dataset, on a computer using an intel(R) i7-8700 CPU with 12 parallel threads. The Monoterpenoides dataset has 286 graphs unevenly divided in 8 classes of at least 10 graphs. Both nodes and egdes are labeled, and the average order is 11.003. Edit costs were set to $c_{vs} = c_{es} = 1 $ and  $c_{vi} = c_{ei} = c_{vr} = c_{er} = 3 $.

Remember that, in a first phase, the proposed algorithm (Sec.~\ref{sec:algo}) identifies a set-median by computing all pairwise distances in the dataset. These distances are computed through two heuristics: \texttt{bipartite} \cite{riebook}, and \texttt{IPFP} \cite{BOUGLEUX2016}. In a second phase, the algorithm iterates the update of a triplet $(\bar{\varphi},\bar{A},\bar{\Phi})$ according to Eq.~\ref{eq:svse} (\textit{i.e.} for vertices either Eq.~\ref{eq:labvert} for Monoterpenoides or Eq.~\ref{eq:eucvert} for Letter, and for edges, Eq.\ref{eq:optphiijlab}--\ref{eq:optaijlab} for Monoterpenoides or Eq.~\ref{eq:optaijunlab} for Letter), and the update of the transformations $\bar{\pi}_p$ using either \texttt{bipartite} or \texttt{IPFP}. We denote by  $m$\texttt{Bipartite} and $m$\texttt{IPFP}  the multistart counterparts of \texttt{Bipartite}, and \texttt{IPFP} \cite{multistart}, where the number of randomly generated initializations was set to 40.

Table \ref{tab:SOD} sums up our results regarding SOD. In Letter and Monoterpenoides, respectively 50 and 10 graphs were picked randomly in each class, and each experiment was repeated 50 times. The results presented in Table \ref{tab:SOD} represent the averages over all classes and all experiments. The four columns SOD SM, t(SM), SOD GM and t(GM) list the SODs and computation times in seconds for the set-median (SM), and the generalized median (GM). Note that t(GM) refers to the computation time of the second phase only.
Using state of the art GED heuristics and making the most of the computed  transformations $\bar{\pi}_p$ to efficiently perform the descent (conversely to many other approaches which use GED only to evaluate candidate medians, without using the detailed transformations), our algorithm produces median graphs with SODs much lower than the set-medians' with a very low running time.
It is noteworthy that the time dedicated to identify the set-median (first phase) is systematically higher than the one dedicated to the generalized median (second phase). Indeed, $|\mathcal{G}|^2$ distances must be computed in the first phase, while $p|\mathcal{G}|$ distances are computed in the second phase, where $p$ denotes the number of iterations before convergence. In practice, we verified that, in most cases, $p<2$ on the letter dataset, and $p<7$ on Monoterpenoides. Interestingly enough, in the hybrid versions of the algorithm (using \texttt{Bipartite} in the first phase and \texttt{IPFP} in the second phase), the alternate descent still produces median graph with reasonably low SOD while starting from a set-median of lesser quality (\emph{i.e.} with higher SODs).

\begin{table}[!t]
\centering
\begin{small}
\resizebox{1\textwidth}{!}{
\begin{tabular}{|c | c | c | c |c | c | c | c | c | c |}
\hline
\multicolumn{2}{|c|}{}&  \multicolumn{4}{c|}{} & \multicolumn{4}{c|}{}  \\[-1em]
\multicolumn{2}{|c|}{Algorithms} & \multicolumn{4}{c|}{Letter (HIGH)} &
\multicolumn{4}{c|}{Monoterpenoides}   \\[3pt]
    1st phase & 2nd phase & SOD SM & t(SM) & SOD GM & t(GM) & SOD SM & t(SM) & SOD GM & t(GM) \\[3pt]
    \hline
  & & & & & & & & &  \\[-1em]
\texttt{Bipartite} & \texttt{Bipartite} & 142.69 & 0.01 & 87.80 & $6*10^{-4}$ & 402.50 &0.002 &253.11 & $8*10^{-4}$ \\[3pt]
 \texttt{Bipartite} & \texttt{IPFP} & 142.87 & 0.013 & 87.61 & 0.003 & 398.01 &0.002 &128.45 &0.179  \\[3pt]
\texttt{IPFP} & \texttt{IPFP} & 135.99 & 0.057 & 87.22 & 0.003 & 202.75 &0.162 &104.11 &0.136  \\[3pt]
$m$\texttt{Bipartite} &$m$\texttt{Bipartite} & 142.04 & 0.014 & 89.47 & $9*10^{-4}$ & 283.94 &0.027 &186.15 &0.01  \\[3pt]
$m$\texttt{Bipartite} & $m$\texttt{IPFP} & 142.19 & 0.018 & 87.66 & 0.013 & 281.14 &0.031 &83.11 &0.545  \\[3pt]
$m$\texttt{IPFP} & $m$\texttt{IPFP} & 135.99 & 0.274 & 87.23 & 0.015 & 106.10 &1.159 &75.08 &0.288  \\[3pt]

\cline{1-10}
\multicolumn{10}{c}

\end{tabular}
}
\end{small}
\caption{SOD computed using different GED approximations.}\label{tab:SOD}
\end{table}

Finally, note that the range between best and worst computed SODs is particularily low on the Letter dataset, while it is rather high on the Monoterpenoides dataset. This seems to indicate that approximate computed distances are close to the optimum in Letter, and far from it in Monoterpenoides.

Picking random trainsets in each class 10\% and 30\% the size of the class, set-medians and generalized medians were computed for each class, and the classification accuracy of a 1-nn algorithm \cite{ferrer-pdh} was evaluated using as training examples: (SM) only the set-median, (GM) only the generalized medians and finally (TS) the whole trainset. Each experiment was repeated 50 times, and Table \ref{tab:classif} presents our results, giving the average preprocessing time \emph{pt} (\emph{i.e.} the time spent in computation of set-medians and generalized medians), as well as classification precisions (denoted by \%) and times for all three training examples considered. Note that the GED heuristic used in the second phase of the algorithms were also used in computing distances by the classifier.

Let us note that our approach competes with a 1-nn classification over the whole trainset, especially when all the distances are computed with a more precise heuristic, such as $m$\texttt{IPFP}. Whenever a precise heuristic is used to compute it, the generalized median appears as a better representative of the class than the set-median. Obviously, classification times are much faster using only the median graphs as training example. 

In few cases, the classification accuracy enabled by set-medians is higher than that enabled by generalized medians. This only happens in cases where computed distances and edit-paths are looser approximations, \emph{i.e.} this always happens on the Monoterpenoides dataset with the $m$\texttt{Bipartite} heuristic used in the initialization phase.

\begin{table}[!t]
\centering
\begin{small}
\begin{tabular}{| c | c | c | c |c | c | c | c | c | c |}
\multicolumn{10}{c}{Letter (HIGH) Dataset} \\[3pt]
 \hline
 & & & & & & & & &  \\[-1em]
 TS & 1st phase & 2nd phase & pt & \% SM & t(SM) & \% GM & t(GM) & \% TS & t(TS) \\
\hline
 & & & & & & & & & \\[-1em]
\multirow{3}{*}{10\%} & $m$\texttt{Bipartite}& $m$\texttt{Bipartite} & 0.023 &76.42 &0.325 &82.82 &0.325 &83.01 &5.275 \\[3pt]
& $m$\texttt{Bipartite}& $m$\texttt{IPFP} & 0.195 &77.40 &5.857 &84.16 &5.771 &83.30 &110.48 \\[3pt]
 &$m$\texttt{IPFP} &$m$\texttt{IPFP}  & 0.447 &78.24 &5.951 &84.60 &5.801 &82.95 &111.84 \\[3pt]
 \hline
 & & & & & & & &  \\[-1em]
 \multirow{3}{*}{30\%}&$m$\texttt{Bipartite} & $m$\texttt{Bipartite}& 0.181 &79.94 &0.251 &84.24 &0.250 &87.24 &11.44 \\[3pt]
 & $m$\texttt{Bipartite}&  $m$\texttt{IPFP} & 0.878 &81.83 &4.323 &86.06 &4.234 &86.86 &239.14 \\[3pt]
  &$m$\texttt{IPFP}&$m$\texttt{IPFP}  &  3.437 &81.59 &4.316 &86.08 &4.245 &86.86 &240.96 \\[3pt]
\cline{1-10}
  \multicolumn{10}{c}{}\\
\multicolumn{10}{c}{Monoterpenoides Dataset} \\[3pt]
\hline
 & & & & & & & &  \\[-1em]
 TS & 1st phase & 2nd phase & pt & \% SM & t(SM) & \% GM & t(GM) & \% TS & t(TS) \\

\hline
 & & & & & & & & & \\[-1em]
 
\multirow{3}{*}{10\%}& $m$\texttt{Bipartite} &$m$\texttt{Bipartite} & 0.054 &32 &0.984 &29.44 &0.957 &51.86 &3.830 \\[3pt]
&$m$\texttt{Bipartite} & $m$\texttt{IPFP} &  1.586 &53.38 &47.96 &57.49 &51.03 &60.69 &186.85 \\[3pt]
 &$m$\texttt{IPFP}  &$m$\texttt{IPFP}  & 2.044 &54.06 &47.31 &62.38 &48.01 &60.69 &187.83 \\[3pt]
 \hline
 & & & & & & & & &  \\[-1em]
\multirow{3}{*}{30\%}&$m$\texttt{Bipartite} & $m$\texttt{Bipartite} & 0.373 &36.39 &0.747 &34.28 &0.732 &67.92 &8.571 \\[3pt]
&$m$\texttt{Bipartite}  &$m$\texttt{IPFP} & 5.148 &54.06 &36.54 &67.79 &37.07 &75.82 &419.81 \\[3pt]
 & $m$\texttt{IPFP} & $m$\texttt{IPFP}  &  15.38 &58.37 &36.15 &74.12 &36.57 &75.94 &419.31 \\[3pt]

\cline{1-10}
\multicolumn{10}{c}{}\\

\end{tabular}
\end{small}
\caption{Classification Results for Letter(HIGH) and Monoterpenoides datasets}\label{tab:classif}
\end{table}

\section{Conclusion}
 We proposed an innovative general method to compute the generalized median graph based on an alternate gradient descent. We showed its efficiency through experiments on two datasets using different edit-cost structures. Computed graphs have much lower SODs than set-medians, and can efficiently be used as representatives in a clustering framework. Quality of computed graph median increases when using accurate rather than fast GED approximation algorithms as sub-routines, especially in the alternate descent phase, while the initialization phase may use different GED heuristics to reach different time/quality compromises. Future developments regarding this promising method include the extension to new edit-cost structures, as well as the possibility to modify the order of the median graph during the optimization process.

\paragraph*{Acknowledgments.}This work is supported by Région Normandie through RIN AGAC project.

\bibliographystyle{plain}
\bibliography{median}

\end{document}